# Pedestrian Behavior Interacting with Autonomous Vehicles: Role of AV Operation and Signal Indication and Roadway Infrastructure

Fengjiao Zou*  Jennifer Ogle†  Weimin Jin‡  Patrick Gerard §  Daniel Petty ¶  Andrew Robb ‖
Clemson University  Clemson University  Arcadis U.S., Inc.  Clemson University  6D Systems  Clemson University

**ABSTRACT**

Interacting with pedestrians is challenging for Autonomous vehicles (AVs). This study evaluates how AV operations /associated signaling and roadway infrastructure affect pedestrian behavior in virtual reality. AVs were designed with different operations and signal indications, including negotiating with no signal, negotiating with a yellow signal, and yellow/blue negotiating/no-yield indications. Results show that AV signal significantly impacts pedestrians' accepted gap, walking time, and waiting time. Pedestrians chose the largest open gap between cars with AV showing no signal, and had the slowest crossing speed with AV showing a yellow signal indication. Roadway infrastructure affects pedestrian walking time and waiting time.

**Index Terms**: Human-centered computing—Human computer interaction (HCI)—Interaction paradigms—Virtual reality

## 1. Introduction

Autonomous vehicles (AVs) are growing in prominence and importance. But some researchers are concerned that pedestrian perception of risk with autonomous vehicles will wane because AVs are programmed to be conservative and will stop for them, even at midblock locations where no crosswalk presents [1]. Such AV risk-averse operations may encourage jaywalking, reduce the transportation system efficiency, and ultimately negatively affect AV receptivity. Therefore, it's necessary to understand variations in pedestrian behavior given different AV operations.

Considering the cost of building a real AV and the safety concerns of having participants cross in front of a test AV in the real world, an increasing number of studies employ virtual reality (VR) to study pedestrian behavior interacting with AVs. Most studies focused on AV communication and intent display (e.g., audio and visual) [2]. Overall, the communication provided by AVs has mostly been proven to help pedestrians make decisions faster [3], [4]. AV operations interacting with pedestrians have also been studied, but most are designed relatively simply; AV negotiation operation was only present in limited research [5], [6]. Some researchers explored roadway infrastructure factors less so compared to the vehicle factors mentioned above [7]. Further, most studies were conducted on two-lane roads, but rarely on multilane roads [6]. Given the typical lane configuration, variation of roadway factors such as median type was rare in prior studies.

In this VR application, we explored how pedestrians' crossing behavior is associated with different AV operations, signals, and roadway infrastructures. The contribution of this paper includes: first, AVs in this study were designed to negotiate with pedestrians in real time with different signal indications; second, it studies the impact of different types of roadway infrastructure on pedestrian and AV interaction; last but not least, pedestrians can interact with multiple AVs in the experiment.

* e-mail: fengjiz@clemson.edu; † e-mail: ogle@clemson.edu
‡ email: Weimin.Jin@arcadis.com; § email: pgerard@clemson.edu
¶ email: daniel@6dsimulations.com; ‖ email: arobb@clemson.edu

## 2. Experiment Design

The virtual environment was created using Unity (2020.3.19f1). We used Oculus Quest 2 with the guardian boundaries turned off to allow participants to cross the multilane road in VR.

### 2.1 AV Operation and Signal design

This research aims to design AV operations to negotiate with pedestrians rather than always yielding to pedestrians at midblock locations. In this study, three AV operations /associated signaling were designed, including AV with negotiation operations but showing no signal (No Signal), AV with negotiation operation showing yellow signal (Y Signal), and AV with negotiation operation showing yellow signal mixed with platoons of non-stopping AVs showing blue signal (YB Signal).

For the Y Signal, once an AV detects a pedestrian and a trajectory conflict between an AV and a pedestrian exits, AV starts negotiating the right of way with pedestrians showing a yellow signal. AVs only yield if pedestrians step into the roadway; otherwise, AVs continue at the current speed. AVs make decisions based on pedestrians' decisions (yield/ not yield) at each timestamp. No Signal has the same AV operation as Y Signal and is designed to compare how AV signal changes pedestrian behavior. A third AV operation/signal scenario is the YB Signal. The blue signal meaning non-stopping at midblock, is a more aggressive operation designed for traffic operation consideration at midblock locations to increase system efficiency. After the platoon of non-stopping AVs, there will be vehicles showing yellow signals coming.

### 2.2 Roadway Infrastructure Design

Research shows that multilane road has a high pedestrian crash rate [8]. This paper designs three multilane roadway infrastructures, including 1) a 4-lane, 2) a 4-lane road with a two-way left-turn lane (abbreviated TWLTL), and 3) a 4-lane road with a median (Median). All lanes are 11 feet wide [9].

### 2.3 Traffic and Gap Design

Gap is "the time lag between two vehicles in any lane encroaching on the pedestrian's crossing path" [9]. This research aims to design gaps that not everyone will cross but also not no one can cross. Based on pilot tests, 4.5 seconds was adopted as the gap between vehicles for the first lane. Vehicle speed is also crucial for pedestrian behavior. After several pilot testing of different AV speeds, the researchers chose the vehicle's speed limit of 20mph. This is a decision made after considering the visibility of the signals from the VR headset and the comfort level of pedestrians crossing.

### 2.4 Experiment Procedure and Data Collection

Training sessions were provided to help participants understand AV signals. Then participants completed three blocks, each containing nine randomized trials that presented combinations of three roadway infrastructures and three AV signals.

Fifty people not pre-disposed to motion sickness participated in this study, most under 30 and four above 60 years old. All but two participants completed each of the 27 trials. A total of 1332 completed trials were used in this analysis. We examined pedestrian accepted gap size, walking time, and waiting time for the first lane in this study. The accepted gap is the gap pedestrians accept to cross the first lane. Walking time is the time pedestrians spend crossing the first lane. And waiting time is the time pedestrians spend waiting at the curb.

For AV signal, section 2.1 defines No, Y, and YB Signal scenarios. To distinguish the yellow signal in Y and YB scenarios, Y_Y was used for yellow from the Y Signal, and Y_YB was used for yellow from YB Signal. B_YB Signal is the blue signal.

## 3. RESULTS AND DISCUSSIONS

This paper used a linear mixed model to explore the association between the accepted gap, walking time, waiting time, and explanatory variables (i.e., roadway infrastructure and AV signal). Overall, AV operations and signals significantly affect pedestrians' behavior, including the accepted gap, walking time, and waiting time. But roadway infrastructure only significantly affects pedestrians walking time and waiting time, not the accepted gap. Besides, no interaction effects were found between the roadway infrastructure and AV signal. Table 1 shows the pairwise differences of AV signals and roadway infrastructures.

Table 1 Pairwise Differences of AV Signals and Roadway Infrastructures

|  | Accepted Gap | | Walking time | | Waiting time | |
| --- | --- | --- | --- | --- | --- | --- |
|  | mean | p. value | mean | p. value | mean | p. value |
| Pairwise differences of AV Signals | | | | | | |
| No - B_YB | 0.409 | <.001* | 0.049 | 0.332 | 1.731 | <.001* |
| Y_Y - B_YB | 0.088 | 0.186 | 0.312 | <.001* | -0.079 | 0.841 |
| Y_YB - B_YB | 0.279 | <.001* | 0.159 | 0.006* | 2.952 | <.001* |
| [1]No - Y_Y | 0.321 | <.001* | -0.263 | <.001* | 1.810 | <.001* |
| No - Y_YB | 0.129 | 0.032* | -0.110 | 0.016* | -1.221 | <.001* |
| [2]Y_Y - Y_YB | -0.191 | 0.002* | 0.153 | <.001* | -3.030 | <.001* |
| Pairwise differences of Roadway Infrastructures | | | | | | |
| 4-lane - Median | -0.077 | 0.130 | 0.129 | <.001* | 1.053 | <.001* |
| 4-lane - TWLTL | -0.068 | 0.178 | 0.109 | 0.004* | 0.654 | 0.029* |
| Median - TWLTL | 0.009 | 0.866 | -0.020 | 0.593 | -0.398 | 0.182 |

* Significant at 0.05 level

Note 1 in Table 1 shows that compared to Y_Y Signal, participants in No Signal accepted 0.321 seconds larger gap, had 0.263 seconds shorter walking time for the first lane, and waited 1.810 seconds longer at the curb, all at a 0.05 significance level. The comparison of No and Y_Y Signal reveals that with the same AV operation (negotiation algorithm), when AV communicates with pedestrians with a yellow signal indication, pedestrians become more confident crossing the road, which complies with the prior study [10]. Note 2 shows that compared to Y_YB, participants in Y_Y accepted a smaller gap, walked slower, and waited less. The results mean that even AVs have the same operation (negotiation with a yellow indicator in this case), pedestrians behave differently depending on the vehicle platoon factor or traffic factor. Pedestrians waited more with the inclusion of blue because most pedestrians will wait until the aggressive blue vehicles pass.

Overall, the largest accepted gap is in No Signal, meaning that when there is no signal indication between AV and pedestrians, pedestrians tend to accept a larger gap because of the uncertain AV operation perceived. The smallest accepted gaps are in Y_Y and B_YB, which are not significantly different. However, pedestrians do walk much faster in the B_YB compared to Y_Y. Further, Y_Y has the longest walking time, indicating that pedestrians take their time walking in front AVs with a yellow indication. The shortest walking time is B_YB and No Signal, the two scenarios where pedestrians perceive the most unsafe and uncertain AV operations. Y_YB has the longest waiting time because most individuals who did not accept the non-stopping AV platoon with a blue signal waited for the first AV with a yellow signal to come. B_YB and Y_Y have the lowest waiting time, which are not significantly different. Those who cross in front of AV with blue signals present pedestrians' risk-taking behavior. And the shortest Y_Y waiting time implies pedestrians immediately step out in front of yellow when they see them because of their trust in AV.

Results show that the roadway infrastructure does not affect the accepted gap, but AV signals do, meaning that the pedestrians' observation of the impending interaction with AVs drives the gap acceptance, not the type of road that pedestrians are negotiating on. However, the infrastructure does affect the walking time and the waiting time. 4-lane undivided road had the longest walking time and waiting time, and median had the shortest. As pedestrians look for gaps across all four lanes, their waiting time increases because there is no intermediate stopping point. The walking time also increases on 4-lane undivided because pedestrians are continuously scanning the environment and engaging decision making, which likely slows them down.

## 4. CONCLUSION

Some scenarios could be challenging for AVs, like the urban area unmarked midblock locations on multilane roads where pedestrians may jaywalk. AVs need to interact with those pedestrians, understand pedestrians' behavioral responses to AV operational and communication strategies. This paper studied pedestrians' unmarked midblock crossing behavior on 4-lane roads (undivided, with a median, or with a TWLTL). Different AV operations and signals were designed in the VR simulation. We conclude that this VR method is useful for studying pedestrian behavior interacting with different AV operations on different roadway infrastructures.

## ACKNOWLEDGEMENTS

We thank Dr. Ronnie Chowdhury for funding the data collection.